# ISIC 2018-A Method for Lesion Segmentation

Hongdiao Wen, Rongjian Xu, Tie Zhang
uestc_wen@outlook.com, andyxusc@163.com, 841310406@qq.com

## Abstract

Our team participate in the challenge of Task 1: Lesion Boundary Segmentation , and use a combined network, one of which is designed by ourselves named updcnn net and another is an improved VGG 16-layer net. Updcnn net uses reduced size images for training, and VGG 16-layer net utilizes large size images. Image enhancement is used to get a richer data set. We use boxes in the VGG 16-layer net network for local attention regularization to fine-tune the loss function, which can increase the number of training data, and also make the model more robust. In the test, the model is used for joint testing and achieves good results.

## Preprocessing

In order to make the limited data more effective, we use image enhancements to enrich the data set[1][2]: flip left and right or up and down, change brightness and contrast, add some noise, resize and crop, etc. These operations can make the data more diverse. In the neural network, it may think that the content of these figures is different, although we humans can easily see that these pictures represent the same patient. But the method of Image enhancement, for the network, it makes sense.

Convolutional neural networks are insensitive to position changes in a picture. Since the conditions of image shooting are limited by the reality, they tend to be relatively simple, but the situations of the test set are difficult to predict. Therefore, image enhancement of the training set can cope with a wider test set. The result of the neural network largely depends on the quality of the input data. We except that the network can understand the image like our human. Through the image enhancement, the neural network can be prevented to learn an unrelated information, so the performance of the neural network is improved. And we also should avoid adding irrelevant data.

## Model

In this challenge, we used the scanet and updcnn networks for joint training and testing.

1. Scanet

Due to the limitations of hardware conditions and the insufficiency of images, in practical applications, we can not obtain a satisfactory result when the amount of data is relatively small. Therefore, using a model that has been trained on a large data set, using the trained parameters, and then using our data to fine-tune the model, is necessary. By using existing models trained with a large data set, it is possible to solve the problem without enough training data and it can also reduce the training time of the model. In the process of solving the task: Lesion Boundary Segmentation, we use the VGG[3] 16-layer net model. Before training, each pixel in the picture will minus the mean of the image. We use 3*3 kernel. It is the smallest unit which can represent different directions. In order to semantically segment the image, we removed the fully connected layer and add several convolutional layers to further handle the feature map generated by VGG 16-layer net.

In order to obtain a better result, we also consider the attention model. When a person sees a scene, he will not notice all the objects in the scene at once, but focus on different objects from primary to secondary, then get a visual understanding of the whole scene by integrating the information. Our brain will let the things that people might interested stay in the center of our view and finally only deal with this part. These research inspire us to use the local attention model to do regularization. According to the image of the training set and ground truth, each picture randomly generates 12 boxes during training. If the share of target exceeds 0.7 in this box, the whole picture is recorded as a positive sample with a target, if the share of target is smaller than 0.1, the entire box is regarded as a background. Do convolution in the area where the box is drawn, and the result is used to calculate the loss function. Put the result into the total loss function as a regularization term. However, we don't give too much weight on the regularization terms, and appropriate adjustments should be made to obtain better results. Using this random-sized box to partition the image, this random processing prevents the model from overfitting. Using the box needs some kind of the user's prior knowledge. When the share is greater than 0.7, the box is considered to include the target in the entire region, and when the share is less than 0.1, the entire region is regarded as a background. If a picture is almost full of one thing, people will ignore other things.

The feature map is obtained by the whole picture and the box need to do some extra convolution after the VGG 16-layer net model and then we have a heatmap. We do up-sampling on the heatmap, so the heatmap can have the same size of the original graph. When do up-sampling, we use simple bilinear interpolation. This simple interpolation method restores the result to the original image size and can achieve good results.

2. Updcnn

Except to use scanet network, we also designed a new network (updcnn) to combining . This network contains a total of 24 convolutional layers, which 22 are directly connected from top to bottom. Considering the degradation phenomenon may occur when the depth of network directly increase, which makes the accuracy saturated and don't increase (not overfitting), we adopt the same idea from ResNet[4]. In addition to directly increasing the depth of network, we let the features generated by the network in middle layers skip Several layers of the network and pass into the later network as input. Considering the problem that the feature scale is not the same, we do not directly pass in, but do a transpose convolution on the feature to ensure that the feature dimension is the same as the feature dimension that needs to be passed in later. The specific model is shown Figure 1.

In order to reduce the number of parameters, in our model, all of our convolution kernels use a size of 3*3, which limits the size of the convolution kernel receptive field, but the multi-layer convolution is actually equal to a layer of convolution kernels with large receptive fields, and each layer will have a relu activation function after convolution. After many layers, it will also increase the nonlinear representation of the model. In addition to using this method to indirectly obtain large receptive fields, we also use the method of dilated convolution to directly obtain the large receptive field of the convolution kernel. Since the convolution kernel we used is only 3*3, so the sampling rate of dilated convolution is only 2, which is equivalent to obtaining a receptive field as a 5*5 convolution kernel size, but the kernel does not have so many parameters.

In our updcnn model, all convolutional layers is followed by a relu activation function . During the uninterrupted convolution layers, all feature sizes are the same. Due to our padding=0 operation when do convolution, the feature will only change in depth dimension, and it will not change in scale. Therefore, all convolution will not affect the size of the scale. There is no pooling operation in our model, which is one the advantage of using the dilated convolution method. The meaning of the pooling operation is to compress the characteristics in the output of the convolutional layer, and to aggregate the features of a small area. It can increase the receptive field while reducing the risk of over-fitting. This operation will make the size of the feature smaller, and it will lose a small part of the feature while refining the feature. We complete this task indirectly using dilated convolution: the interval sampling of the dilated convolution allows the feature to complete the pooling task while doing convolution.

Our model has a total depth of 22 layers. In addition to these layers, 2 layers of additional transposed convolutional layers are added to it as a link. The size of the convolution kernel used in the whole network is 3*3, and each convolution layer contains an activation function (relu). Every three convolutional layers are regarded as a group, after each group, a deconvolution layer or an adding operation is followed, used to scale the entire feature map or combine two sets of deconvolution features to the next convolutional layer. The entire network has 6 sets of convolutional layers . The filter number of each convolution layer is different. The first convolution layer has 64 filters, after that, each layer has 32 more layers than the previous layer, that is, the second is 96, and the third is 128, until 224, two sets of convolutional layers. After that, a deconvolution layer is directly followed. The number of the filter in deconvolution layer is 256, which is the maximum value of the number of filter in the convolution layer. After that, the connected convolution layer is changed to 64 layers again, and is increased according to the above rule. After three such steps, a convolution layer of 2 filters is generated to produce a characteristic map of the output. In total, there are three deconvolution layers directly connected to the entire network, and there is no pooling layer in the network. Each deconvolution layer will double the scale of the feature map, so the scale of the output feature map is 8 times as large as the input image.

## Training and experiment

The architecture of our network is made up of the scanet network and the updcnn network. The result is stacked by the output of the two network. Therefore, when training the network, we will train the two networks together. In order to improve the correct rate of the results, we first train the two networks separately, and the parameters generated are used as the pre-training values for the joint training. When we train each network separately, since the scanet network is an improvement of vgg16 network, we use the parameter value trained by vgg16 on the imagenet as its pre-trained value. The updcnn network is designed by us. There is no pre-training value. We start training directly. When the two parts of the network have been trained, we use the parameters generated by the individual training as the pre-training values of the joint training.

In the training set and verification set, the image size is processed to 512*512, and the input sizes of our two networks are different. The input size of the scanet is 320*320, and the input size of updcnn is 40*40. After doing the image enhancements, we linearly reduce the scale of the image to 320 and 40. Since the given image is square, there is no need to crop the image. The feature image produced by scanet will not produce scale changes, and updcnn will enlarge the scale of the input image by 8 times, and finally get two identical size maps. We will superimpose the feature maps generated by the two networks to obtain multi-channels feature image. After obtaining the multi-channel feature image, we reserve the maximum value of each pixel to get a single-channel image. This image is the feature image we ultimately produced. We binarize the annotation image of the training set, setting the lesion area to 1 and the non-pathological area to 0, compared with our feature map to generate a loss function. And the regularization term generated by the box is also added to the loss function.

In the test, the threshold value is used to determine the pixel class of the feature map, and tune threshold can increase the adaptability of the model to the data set. The test results are shown in Figure 2.

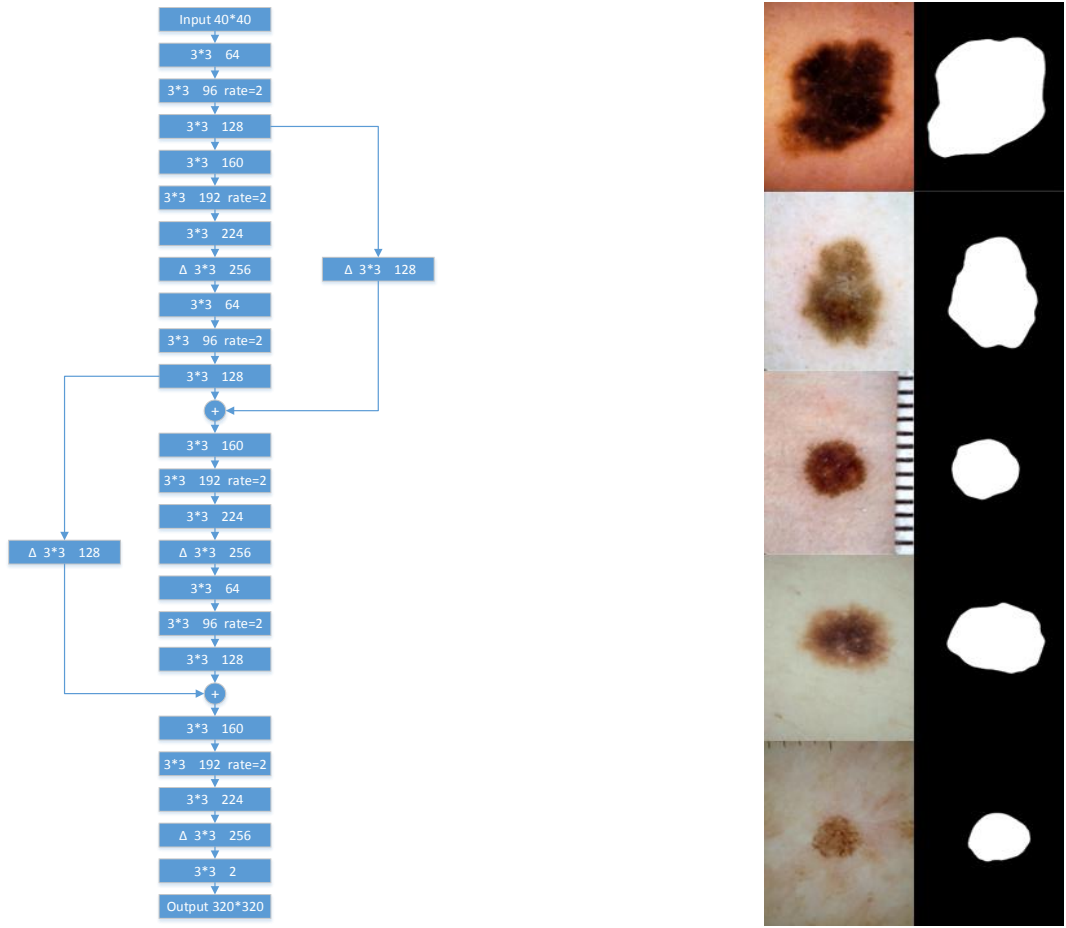

Figure 1 updcnn, Δ means dilated convolution                    Figure 2 result